\documentclass[10pt,twocolumn,letterpaper]{article}

\usepackage{iccv}              
\PassOptionsToPackage{dvipsnames,table}{xcolor} 
\definecolor{iccvblue}{rgb}{0.21,0.49,0.74}
\usepackage[pagebackref,breaklinks,colorlinks,allcolors=iccvblue]{hyperref}
\usepackage{amsmath} 
\usepackage{amsfonts} 
\usepackage{bm} 

\usepackage{caption}
\usepackage{amssymb}
\definecolor{realtimecolor}{rgb}{1.,1,0.85}
\definecolor{CheckGreen}{rgb}{0, 0.55, 0}
\definecolor{XRed}{RGB}{180,0,0}
\usepackage{multirow}
\usepackage{eucal}
\usepackage{amssymb}
\usepackage{pifont}
\newcommand{\cmark}{\ding{51}}%
\newcommand{\xmark}{\ding{55}}%
\usepackage{cuted}
\AfterEndEnvironment{strip}{\leavevmode}

\DeclareMathOperator*{\argmin}{arg\,min}

\newtoggle{comments}
\toggletrue{comments}
\iftoggle{comments}{%
    \newcommand{\tarasha}[1]{{\leavevmode\color{magenta}[Tarasha: #1]}}
    \newcommand{\deva}[1]{{\leavevmode\color{blue}[Deva: #1]}}
    \newcommand{\neehar}[1]{{\leavevmode\color{red}[Neehar: #1]}}
    \newcommand{\jeff}[1]{{\leavevmode\color{green}[Jeff: #1]}}
    \newcommand{\zihan}[1]{{\leavevmode\color{orange}[Zihan: #1]}}
}{%
  \newcommand{\tarasha}[1]{}
  \newcommand{\deva}[1]{}
  \newcommand{\neehar}[1]{}
  \newcommand{\jeff}[1]{}
  \newcommand{\zihan}[1]{}
}

\def\paperID{111} 
\def\confName{ICCV}
\def\confYear{2025}

\title{MonoFusion: Sparse-View 4D Reconstruction via Monocular Fusion}

\author{Zihan Wang, Jeff Tan, Tarasha Khurana\thanks{Equal Contribution.}, Neehar Peri$^*$, Deva Ramanan \\
Carnegie Mellon University
}

\begin{document}

\twocolumn[{%
\renewcommand\twocolumn[1][]{#1}%
\maketitle
\begin{center}
\includegraphics[width=0.95\linewidth]{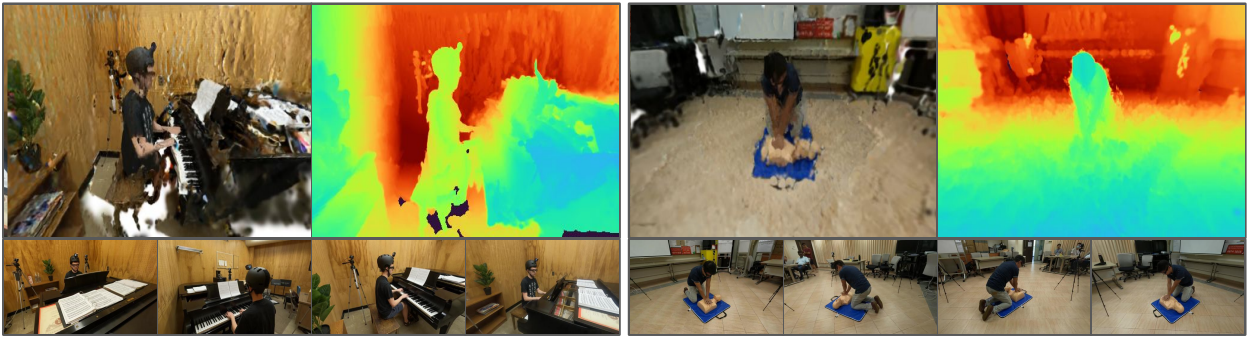}
\captionof{figure}{\textbf{Dynamic Scene Reconstruction from Sparse Views}. MonoFusion reconstructs dynamic human behaviors, such as playing the piano or performing CPR, from four equidistant inward-facing static cameras. We visualize the RGB and depth renderings of a 45$^\circ$ novel view between two training views. Training views are shown below for reference.
}
\label{fig:teaser}
\end{center}%
}]

\begin{abstract}
We address the problem of dynamic scene reconstruction from sparse-view videos. Prior work often requires dense multi-view captures with hundreds of calibrated cameras (e.g. Panoptic Studio). Such multi-view setups are prohibitively expensive to build and cannot capture diverse scenes in-the-wild. In contrast, we aim to reconstruct dynamic human behaviors, such as repairing a bike or dancing, from a small set of sparse-view cameras with complete scene coverage (e.g. four equidistant inward-facing static cameras). We find that dense multi-view reconstruction methods struggle to adapt to this sparse-view setup due to limited overlap between viewpoints. To address these limitations, we carefully align independent monocular reconstructions of each camera to produce time- and view-consistent dynamic scene reconstructions. Extensive experiments on PanopticStudio and Ego-Exo4D demonstrate that our method achieves higher quality reconstructions than prior art, particularly when rendering novel views. 


\end{abstract}    
\vspace{-10pt}
\section{Introduction}
\label{sec:intro}

Accurately reconstructing dynamic 3D scenes from multi-view videos is of great interest to the vision community, with applications in AR/VR \cite{song2023total, lin2025camerabench, jiang2025stg, lee2025skyfall, sunmola2025surgical, li2023hong}, autonomous driving \cite{lu2024drivingrecon, lu20254d, lu2025uniugp, pan2024harmonicnerf,zeng2024driving, zeng2025FSDrive}, and robotics \cite{10969987, 10801738}. Prior work often studies this problem in the context of dense multi-view videos, which require dedicated capture studios that are prohibitively expensive to build and are difficult to scale to diverse scenes in-the-wild. In this paper, we aim to strike a balance between the ease and informativeness of multi-view data collection by reconstructing skilled human behaviors (e.g., playing a piano and performing CPR) from four equidistant inward-facing static cameras (Fig. \ref{fig:teaser}).  

\begin{figure}[t!]
    \centering
    \includegraphics[width=\linewidth]{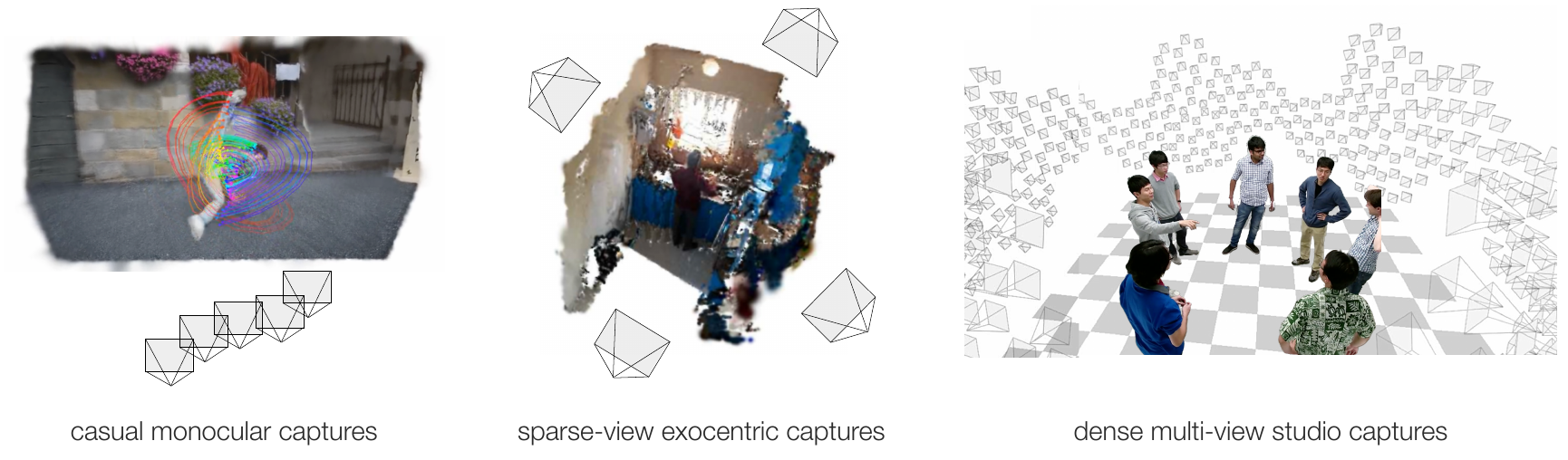}
    \caption{\textbf{Problem Setup.} Our sparse-view setup ({\bf middle}) strikes a balance between ill-posed reconstructions from casual monocular captures \cite{gao2022dynamic, davis} and well-constrained reconstructions from dense multi-view studio captures \cite{joo2017panoptic}. Unlike existing ``sparse-view" datasets like DTU~\cite{jensen2014large} and LLFF~\cite{mildenhall2019llff}, our setup is more challenging because input views are 90$^\circ$ apart with limited cross-view correspondences.
   }
   \label{fig:onecolffff}
\end{figure}

\vspace{-10pt}
\paragraph{Problem setup.} Despite recent advances in dynamic scene reconstruction ~\cite{gao2021dynamic, dynnerf, gao2022dynamic, chen2024survey, lu2025rhythmguassian, liu2025monosplat, chan2026adagar,li2023hong}, current approaches often require dozens of calibrated cameras~\cite{luiten2023dynamic, joo2017panoptic}, are category specific~\cite{yang2022banmo}, or struggle to generate multi-view consistent geometry~\cite{li2023dynibar}. We study the problem of reconstructing dynamic human behaviors from an {\em in-the-wild capture studio}: a small set of (4) portable cameras with limited overlap but complete scene coverage, such as in the large-scale Ego-Exo4D dataset~\cite{grauman2023egoexo4d}. We argue that sparse-view limited-overlap reconstruction presents unique challenges not found in dense multi-view setups and typical ``sparse view" captures with large covisibility (Fig.~\ref{fig:onecolffff}). For dense multi-view captures, it is often sufficient to rely solely on geometric and photometric cues for reconstruction, often making use of classic techniques from (non-rigid) structure from motion~\cite{forsyth2003modern}. As a result, these methods fail in sparse-view settings with limited cross-view correspondences. 


\vspace{-10pt}
\paragraph{Key insights.} We find that initializing sparse-view reconstructions with monocular geometry estimators like  MoGe~\cite{wang2024moge} produces higher quality results. However, naively merging independent monocular geometry estimates often yields inconsistent geometry across views (e.g. duplicate structures), resulting in a local minima during 3D optimization. Instead, we carefully align monocular reconstructions (that are {\em independently} predicted for each view and time) to a global reference frame that is learned from a {\em static} multi-view reconstructor (like DUSt3R~\cite{wang2024dust3r}).
Furthermore, many challenges in inferring view-consistent and time-consistent depth become dramatically simplified when working with {\em fixed cameras with known poses} (inherent to the in-the-wild capture setup that we target). For example, temporally consistent background geometry can be enforced by simply averaging predictions over time. 

\vspace{-10pt}
\paragraph{Contributions.} We present three major contributions.
\begin{itemize}
    \item We highlight the challenge of reconstructing skilled human behaviors in dynamic environments from sparse-view cameras in-the-wild.
    \item We demonstrate that monocular reconstruction methods can be extended to the sparse-view setting by carefully incorporating monocular depth and foundational priors.
    \item We extensively ablate our design choices and show that we achieve state-of-the-art performance on PanopticStudio and challenging sequences from Ego-Exo4D.
\end{itemize}
\section{Related Work}
\label{sec:related work}

\paragraph{Dynamics scene reconstruction.}
Dynamic scene reconstruction~\cite{chen2024survey} has received significant interest in recent years. While classical work~\cite{newcombe2015dynamicfusion, dou2016fusion} often relies on RGB-D sensors, or strong domain knowledge~\cite{carranza2003free,deAguiar-2008}, recent approaches \cite{li2023dynibar} based on neural radiance fields \cite{mildenhall2020nerf} have progressed towards reconstructing dynamic scenes in-the-wild from RGB video alone. However, such methods are computationally heavy, can only reconstruct short video clips with limited dynamic movement, and struggle with extreme novel view synthesis. Recently, 3D Gaussian Splatting \cite{kerbl20233d, luiten2023dynamic} has accelerated radiance field training and rendering via an efficient rasterization process. Follow-up works \cite{yang2023deformable3dgs, wu20234dgaussians, liang2023gaufre} repurpose 3DGS to reconstruct dynamic scenes, often by optimizing a fixed set of Gaussians in canonical space and modeling their motion with deformation fields. However, as \citet{gao2022dynamic} points out, such methods often struggle to reconstruct realistic videos. Many works address this shortcoming by relying on 2D point tracking priors \cite{wang2024som}, fusing Gaussians from many timesteps \cite{lei2024mosca}, modeling isotropic Gaussians \cite{stearns2024dgmarbles}, or exploiting domain knowledge such as human body priors \cite{tan2024dressrecon}. However, these approaches study the reconstruction problem in the monocular setting. 
As 4D reconstruction from a single viewpoint is under-constrained, practical robotics setups for manipulation \cite{khazatsky2024droid} and hand-object interaction \cite{kwon2021h2o, fan2023arctic, wang2024ho} adopt camera rigs where a sparse set of cameras capture the scene of interest. Similarly, datasets like Ego-Exo4D \cite{grauman2023egoexo4d}, DROID \cite{khazatsky2024droid} and H2O \cite{kwon2021h2o} explore sparse-view capture for dynamic scenes in-the-wild.

\vspace{-10pt}
\paragraph{Novel-view synthesis from sparse views.}
Both NeRF and 3D Gaussian Splatting require dense input view coverage, which hinders their real-world applicability. Recent works aim to reduce the number of required input views by adding additional supervision and regularization, such as depth \cite{deng2021depth} or semantics \cite{yu2021pixelnerf}. FSGS~\cite{zhu2024fsgs} builds on Gaussian splatting by producing faithful static geometry from as few as three views by unpooling existing Gaussians and adopting extra depth supervision. Recent studies such as ~\cite{yang2024gaussianobject}, on the other hand, adds noise to Gaussian attributes and relies on a pre-trained ControlNet~\cite{zhang2023controlnet} to repair low-quality rendered images. Other works such as MVSplat~\cite{chen2024mvsplat} build a cost volume representation and predict Gaussian attributes in a feed-forward manner. However, they can only synthesize novel views with small deviations from the nearest training view. For methods that rely on learned priors, high-quality novel view synthesis is often limited to images within the training distribution. Such methods cannot handle diverse real-world geometry. Diffusion-based reconstruction methods~\cite{gao2024cat3d, wu2024reconfusion, zhao2025diffusionsfm} try to generate additional views consistent with the sparse input views, but often produce artifacts. In our case, four sparse view cameras are separated around 90$^\circ$ apart, posing unique challenges.

\begin{figure*}[t]
    \centering
    \includegraphics[width=.85\linewidth]{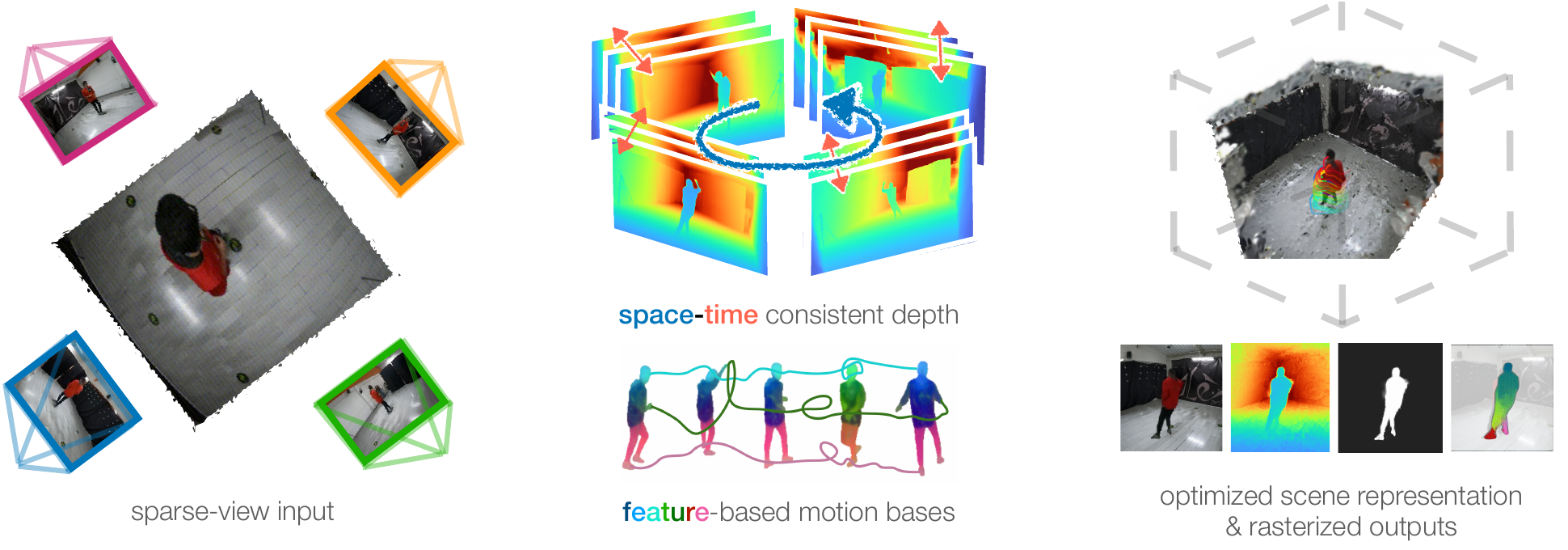}
    \vspace{-10pt}
    \caption{\textbf{Approach.} Given sparse-view video sequences of a scene (left), we aim to optimize a 3D gaussian representation over time. We begin by running DUSt3R \cite{wang2024dust3r}, a {\em static} multi-view reconstruction method, on the sparse views of a given reference timestamp. This generates a global reference frame that connects all views. Next, we use MoGe \cite{wang2024moge} to independently predict depth maps for each camera. Since these depth predictions are only defined up to an {\em affine transformation}, we must estimate a scale and shift for each predicted depth map across all views and time instants. To achieve this, we leverage the fact that background pixels remain static over time. Specifically, for each time instant and each view, we align the background regions of each camera's depth map to the global reference frame by adjusting the scale and shift parameters accordingly (middle, top). This process requires a foreground-background mask for all input videos (which can be obtained using off-the-shelf tools like SAM 2 \cite{ravi2024sam}). To reduce occlusions and noisy depth predictions, we concatenate all aligned background depth points, and average corresponding background points (where correspondence across time is trivially given by the 2D pixel index of the unprojected pointmap) across time. Lastly, we find that motion bases constructed from feature-clustering form a more geometrically consistent set of bases (middle, bottom), than those initialized by noisy 3D tracks \cite{wang2024som}. Our optimization yields a 4D scene representation from which we can rasterize RGB frames, depth maps, a foreground silhouette, and object features from novel views (right).
    }
    \label{fig:method}
\end{figure*}


\vspace{-10pt}
\paragraph{Feed-forward geometry estimation.}
Learning-based methods, such as monocular depth networks, are able to reconstruct 3D objects and scenes by learning strong priors from training data. While early works \cite{eigen2014depth, fu2018deep} focus on in-domain depth estimation, recent works build foundational depth models by scaling up training data \cite{ranftl2021vision, yang2024depthanythingv2, wang2024moge}, resolving metric ambiguity from various camera models \cite{hu2024metric3dv2, piccinelli2024unidepth}, or relying on priors such as Stable Diffusion \cite{rombach2021highresolution, ke2023marigold, fu2024geowizard}. Unfortunately, monocular depth networks are not scale or view consistent, and often require extensive alignment against ground-truth to produce meaningful metric outputs. 
To address these shortcomings, DUSt3R \cite{wang2024dust3r} and MonST3R \cite{zhang2024monst3r} propose the task of point map estimation, which aims to recover scene geometry as well as camera intrinsics and extrinsics given a pair of input images. These methods unify single-view and multi-view geometry estimation, and enable consistent depth estimation across either time or space.
\section{Towards Sparse-View 4D Reconstruction}
Given sparse-view (i.e. 3 -- 4) videos from stationary cameras as input, our method recovers the geometry and motion of a dynamic 3D scene (Fig. \ref{fig:method}). We model the scene as canonical 3D Gaussians (Sec. \ref{sec:preliminaries}), which translate and rotate via a linear combination of motion bases.
We initialize consistent scene geometry by carefully aligning geometry predictions from multiple views (Sec. \ref{sec:consistent_depth}), and initialize motion trajectories by clustering per-point 3D semantic features distilled from 2D foundation models (Sec. \ref{sec:feature_clustering}).
We formulate a joint optimization which simultaneously recovers geometry and motion (Sec. \ref{sec:optimization}).



\subsection{3D Gaussian Scene Representation}
\label{sec:preliminaries}

We represent the geometry and appearance of dynamic 3D scenes using 3D Gaussian Splatting \cite{kerbl20233d}, due to its efficient optimization and rendering. Each Gaussian in the canonical frame $t_0$ is parameterized by $(\mathbf{x}_0,\mathbf{R}_0,\mathbf{s},\alpha,\mathbf{c})$, where $\mathbf{x}_0\in\mathbb{R}^3$ is the Gaussian's position in canonical frame, $\mathbf{R}_0\in\mathbb{SO}(3)$ is the orientation, $\mathbf{s}\in\mathbb{R}^3$ is the scale, $\alpha\in\mathbb{R}$ is the opacity, and $\mathbf{c}\in\mathbb{R}^3$ is the color. The position and orientation are time-dependent, while the scale, opacity, and color are persistent over time. 
We additionally assign a semantic feature $\mathbf{f}\in\mathbb{R}^N$ to each Gaussian (Sec. \ref{sec:feature_clustering}), where $N=32$ is an arbitrary number representing the embedding dimension of the feature. Empirically, we find that fixing the color and opacity of Gaussians results in a better performance. In summary, for the $i$-th 3D Gaussian, the optimizable attributes are given by $\Theta^{(i)} = \{\mathbf{x}^{(i)}_0, \mathbf{R}^{(i)}_0, \mathbf{s}^{(i)}, \mathbf{f}^{(i)} \}$. Following \cite{zhou2024feature}, the optimized Gaussians are rendered from a given camera to produce an RGB image and a feature map using a tile-based rasterization procedure.

\subsection{Space-Time Consistent Depth Initialization}
\label{sec:consistent_depth}

Similar to recent methods \cite{szymanowicz2024splatter, wang2024som}, we rely on data-driven monocular depth priors to initialize the position and appearance of 3D Gaussians over time. Given the success of initializing 3DGS with monocular depth in single-view settings \cite{wang2024som}, one might think to naturally extend this to multi-view settings by independently initializing from monocular depth for each view. However, this yields conflicting geometry signals, as monocular depth estimators commonly predict up to an unknown scale and shift factor. Thus, the unprojected monocular depths from separate views are often inconsistent, resulting in duplicated object parts. 

\vspace{-10pt}
\paragraph{Multi-view pointmap prediction.} DUSt3R \cite{wang2024dust3r} predicts multi-view consistent pointmaps across $K$ input images by first inferring pairwise pointmaps, followed by a global 3D optimization that searches for per-image pointmaps and pairwise similarity transforms
(rotation, translation, and scale) that best aligns all pointmaps with each other.
We run DUST3R on the multiview images at time $t$, but constrain the global optimization to be consistent with the $K$ known stationary camera extrinsics $\{\mathbf{P}_k\}$ and intrinsics $\{\mathbf{K}_k\}$. This produces per-image global pointmaps $\{\chi^t_k\}$ in metric coordinates.
One can then compute a depth map by simply projecting each pointmap back to each image with the known cameras
\begin{equation}
    d^t_k(u,v)\begin{bmatrix}u & v &1\end{bmatrix}^T = \mathbf{K}_k\mathbf{P}_k{\chi}^t_k(u,v)
\end{equation}
This produces metric-scale multi-view consistent depth maps $d^t_k(u,v)$, which are still not consistent over time.



\vspace{-10pt}
\paragraph{Spatio-temporal alignment of monocular depth with multi-view consistent pointmaps.} 
In fact, even beyond temporally inconsistency, such multiview predictors tend to underperform on humans since they are trained on multiview data where dynamic humans are treated as outliers. Instead, we find monocular depth estimators such as MoGe \cite{wang2024moge} to be far more accurate, but such predictions are not metric (since they are accurate only up to an affine transformation) and are not guaranteed to be consistent across views {\em or} times.
Instead, our strategy is to use the multiview depth maps from DUST3R as a metric target to {\em align} monocular depth predictions, which we write as $m^t_k(u,v)$.
Specifically, we search for scale and shift factors $a_k^t$ and $b_k^t$ that minimize the following error:
\vspace{5pt}
\begin{equation}
\argmin_{\{a^t_k,b^t_k\}}\sum_{t=1}^T\sum_{k=1}^K\sum_{u,v \in \text{BG}^t_k}\left\lVert (a^t_km^t_k(u,v) + b^t_k) -d^t_{k}(u,v)\right\rVert^2
\end{equation}

where $\text{BG}^t_k$ refers to a pixelwise background mask for camera $k$ at frame $t$. The above uses metric background points as a target for aligning all monodepth predictions. The above optimization can be solved quite efficiently since each time $t$ and view $k$ can be optimized independently with a simple least-squares solver (implying our approach will easily scale to long videos). However, the above optimization will still produce scale factors that are not temporally consistent since the targets are temporally inconsistent as well. But we can exploit the constraint that background points should be {\em static} across time for stationary cameras. To do so, we replace $d^t_{k}(u,v)$ with a static target $d_k(u,v)$ obtained by averaging depth maps over time or selecting a canonical reference timestamp.
The final set of scaled time- and view-consistent depthmaps are then unprojected back to 3D pointmaps. Note that this tends to produce accurate predictions for static background points, but the dynamic foreground may remain noisy because they cannot be naively denoised by simple temporal averaging. Rather, we rely on motion-based 3DGS optimization to enforce smoothness of the foreground, described next.

During our experiments, we identified two additional limitations that significantly impact visual quality.\\
\emph{(1) Scale initialization}: We observed that initializing 3D Gaussian scales with $k$-nearest neighbors often results in poor appearance, such as extremely large Gaussians filling empty space and blurring the background. To address this, we follow SplaTAM \cite{keetha2024splatamsplattrack} and initialize each Gaussian scale based on its projected pixel area: $\textrm{scale}=\frac{d}{0.5(f_x+f_y)}$, where $d$ is a pixel's depth and $f_x,f_y$ are focal lengths.\\
\emph{(2) Insufficient Gaussian density:} Using only one Gaussian per input pixel fails to adequately capture fine details. We instead initialize 5 Gaussians per input pixel, providing better representation of fine details.

\begin{figure*}[t]    \includegraphics[width=0.97\textwidth]{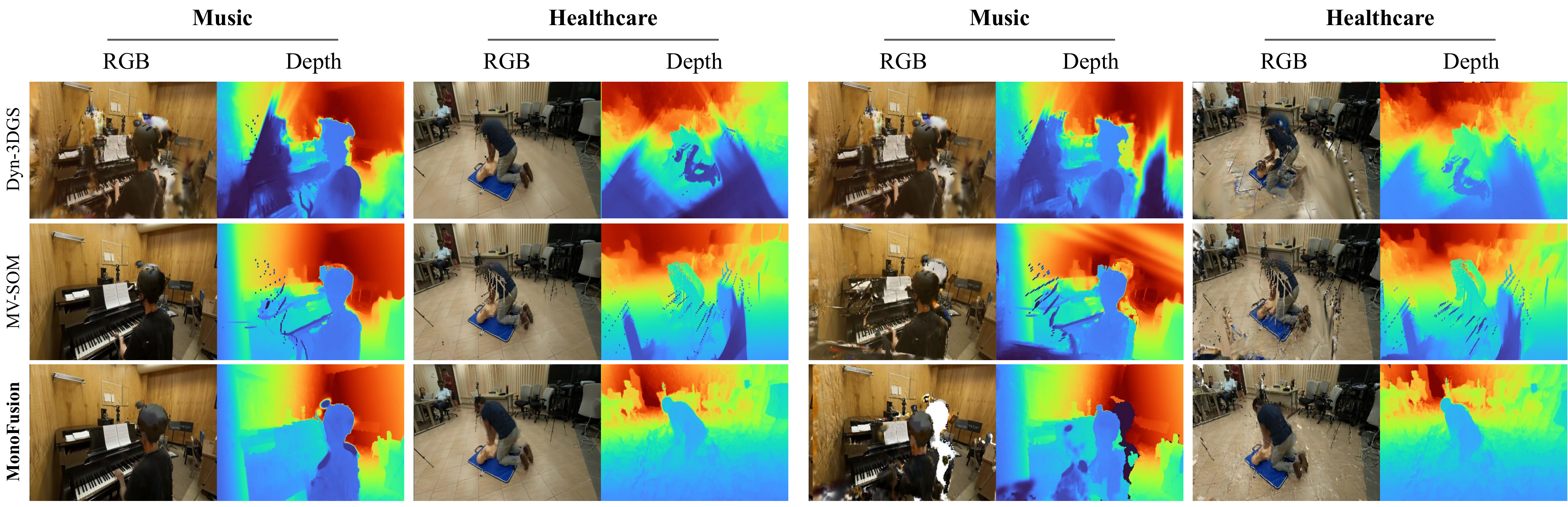}
    \captionof{figure}{\textbf{Qualitative analysis of held-out view synthesis on ExoRecon.} We show qualitative results of held-out view synthesis (\textbf{left}) and a 5$^\circ$ deviation from the static camera position at the held-out timestamp (\textbf{right}). As compared to other multi-view baselines, our method does dramatically better at interpolating the motion of dynamic foreground (left), even from new camera views (right). We posit that Dynamic 3DGS suffers because of lack of geometric constraints and MV-SOM has duplicate foreground artifacts because of conflicting depth initialization from the four views. 
    }
    \label{fig:noveltimesynthesis}
\end{figure*}

\subsection{Grouping-based Motion Initialization}
\label{sec:feature_clustering}

Beyond initializing time- and view-consistent geometry in the canonical frame, we also aim to initialize reasonable estimates of the scene motion. We model a dynamic 3D scene as a set of $\mathcal{N}$ canonical 3D Gaussians, along with time-varying rigid transformations $\mathbf{T}_{0\to t}=[\mathbf{R}_{0\to t} \mathbf{t}_{0\to t}]\in\mathbb{SE}(3)$ that warp from canonical space to time $t$:
\begin{equation}
    \mathbf{x}_t=\mathbf{R}_{0\to t}\mathbf{x}_0+\mathbf{t}_{0\to t}\quad \mathbf{R}_t=\mathbf{R}_{0\to t}\mathbf{R}_0
\end{equation}

\vspace{-10pt}
\paragraph{Motion bases.} Similar to Shape of Motion \cite{wang2024som}, we make the observation that in most dynamic scenes, the underlying 3D motion is often low-dimensional, and composed of simpler units of rigid motion. For example, the forearms tend to move together as one rigid unit, despite being composed of thousands of distinct 3D Gaussians. Rather than storing independent 3D motion trajectories for each 3D Gaussian $(i)$, we define a set of $B$ learnable basis trajectories $\{\mathbf{T}^{(i,b)}_{0\to t}\}_{b=1}^B$. The time-varying rigid transforms are written as a weighted combination of basis trajectories, using fixed per-point basis coefficients $\{w^{(i,b)}\}_{b=1}^B$:
\begin{equation}
    \mathbf{T}^{(i)}_{0\to t}=\sum_{b=1}^B\mathbf{w}^{(i,b)}\mathbf{T}_{0\to t}^{(i,b)}
\end{equation}


\vspace{-10pt}
\paragraph{Motion bases via feature clustering.} Unlike Shape of Motion which initializes motion bases by clustering 3D tracks, our key insight is that semantically grouping similar scene parts together can help regularize dynamic scene motion, without ever initializing trajectories from noisy 3D track predictions. Inspired by the success of robust and universal feature descriptors \cite{oquab2023dinov2}, we obtain pixel-level features for each input image by evaluating DINOv2 on an image pyramid. We average features across pyramid levels and reduce the dimension to 32 via PCA \cite{amir2021deep}. We choose the small DINOv2 model with registers, as it produces fewer peaky feature artifacts \cite{darcet2023vision}.

Given the consistent pixel-aligned pointmaps $\chi_{t,k}^{\text{(time+view)}}$, we associate each pointmap with the 32-dim feature map $\mathbf{f}_{t,k}$ computed from the corresponding image. We perform k-means clustering on per-point features $\mathbf{f}$ to produce $b$ initial clusters of 3D points. After initializing 3D Gaussians from pointmaps, we set the motion basis weight $\mathbf{w}^{(i,b)}$ to be the L2 distance between the cluster center and 3D Gaussian center. We initialize the basis trajectories $\mathbf{T}^{(b)}_{0\to t}$ to be identity, and optimize them via differentiable rendering.


\begin{table*}[t]
\centering
\definecolor{region1}{HTML}{E8F4FF} 
\definecolor{region2}{HTML}{FFE6D9} 
\definecolor{region3}{HTML}{E8FFE8} 
\definecolor{highlight}{HTML}{FFD6D6} 
\resizebox{\textwidth}{!}{%
\begin{tabular}{cl|cccc|cccc} 
\toprule
\multirow{2}{*}{Dataset} & \multirow{2}{*}{Method}
& \multicolumn{4}{c}{Full Frame} & \multicolumn{4}{c}{Dynamic Only} \\
\cmidrule(lr){3-6} \cmidrule(lr){7-10} 
& & PSNR $\uparrow$ & SSIM $\uparrow$ & LPIPS $\downarrow$ & AbsRel $\downarrow$  
& PSNR $\uparrow$ & SSIM $\uparrow$ & LPIPS $\downarrow$ & IOU $\uparrow$ \\ 

\midrule
& SOM \cite{wang2024som} & 17.86 & 0.687 & 0.460 &  0.491 & 18.75 & 0.701 &  0.236 & 0.358 \\

& Dyn3D-GS \cite{luiten2023dynamic} &  25.37 & 0.831 & 0.266 &  0.207 & 26.11 & 0.862 & 0.129 &  ---   \\

\multirow{-2.2}{*}{\textbf{Panoptic Studio}} & MV-SOM \cite{wang2024som} & 26.28 &  0.858& 0.241 & 0.331 &  26.80 &  0.883 & 0.161 & 0.886 \\
& \textbf{MonoFusion}  & \textbf{28.01} & \textbf{0.899} & \textbf{0.117} & \textbf{0.149} & \textbf{27.52}  & \textbf{0.944} & \textbf{0.022}  & \textbf{0.965} \\

\midrule
& SOM \cite{wang2024som} & 14.73  & 0.535 & 0.482  & 0.843 & 15.63 &0.559  & 0.450 & 0.294  \\

& Dyn3D-GS \cite{luiten2023dynamic} & 24.28 & 0.692 & 0.539 & 0.612 & 24.61 & 0.673 &  0.384 & ---   \\
& MV-SOM-DS \cite{wang2024som}& 28.37 & 0.906 & 0.079 & 0.398 & 28.23 & 0.931 & 0.063 & 0.872 \\
\multirow{-3}{*}{\textbf{ExoRecon}} & MV-SOM \cite{wang2024som} &  26.91 &  0.890& 0.138  & 0.474  & 27.31  & 0.919 & 0.078 & 0.845 \\
& \textbf{MonoFusion}  & \textbf{30.43} & \textbf{0.927} & \textbf{0.061} & \textbf{0.290} & \textbf{29.71}  & \textbf{0.947} & \textbf{0.017} & \textbf{0.963} \\

\bottomrule
\end{tabular}
}
\caption{\textbf{Quantitative analysis of held-out view synthesis.} We benchmark our method against state-of-the-art approaches by evaluating the novel-view rendering and geometric quality on both the dynamic foreground region and the entire scene, across the held-out frames from input videos.  MV-SOM is a multi-view version of Shape-of-Motion \cite{wang2024som} that we construct by instantiating four different instances of single-view shape of motion, and optimize them together. On Panoptic Studio, groundtruth depth for computing the AbsRel metric is obtained from 27-view optimization of the original Dynamic 3DGS, and for ExoRecon, we project the released point clouds obtained via SLAM from Aria glasses. When evaluating single-view baselines, SOM \cite{wang2024som}, we naively aggregate their predictions from the four views and evaluate this aggregated prediction against the evaluation cameras.
\label{exp:ucsd_numerical}}
\end{table*}


\subsection{Optimization}
\label{sec:optimization}

As observed in prior work \cite{li2021neural, gao2021dynamic}, using photometric supervision alone is insufficient to avoid bad local minima in a sparse-view setting. Our final optimization procedure is a combination of photometric losses, data-driven priors, and regularizations on the learned geometry and motions.

During each training step, we sample a random timestep $t$ and camera $k$. We render the image $\mathbf{\hat I}_{t,k}$, mask $\mathbf{\hat M}_{t,k}$, features $\mathbf{\hat F}_{t,k}$, and depth $\mathbf{\hat D}_{t,k}$. We compute reconstruction loss by comparing to off-the-shelf estimates:
\begin{equation}
\resizebox{\columnwidth}{!}{$
    \mathcal{L}_{\text{recon}}=
    \left\lVert\mathbf{\hat I}-\mathbf{I}\right\rVert_1+
    \lambda_{\text{m}}\left\lVert\mathbf{\hat M}-\mathbf{M}\right\rVert_1+
    \lambda_{\text{f}}\left\lVert\mathbf{\hat F}-\mathbf{F}\right\rVert_1+
    \lambda_{\text{d}}\left\lVert\mathbf{\hat D}-\mathbf{D}\right\rVert_1
$}
\end{equation}

We additionally enforce a rigidity loss between randomly sampled dynamic Gaussians and their $k$ nearest neighbors. Let $\mathbf{\hat X}_t$ denote the location of a 3D Gaussian at time $t$, and let $\mathbf{\hat X}_{t'}$ denote its location at time $t'$. Over neighboring 3D Gaussians $i$, we define:
\begin{equation}
    \mathcal{L}_{\text{rigid}}=\sum_{\text{neighbors }i}\left\lVert\mathbf{\hat X}_t-\mathbf{\hat X}^{(i)}_t\right\rVert_2^2-\left\lVert\mathbf{\hat X}_{t'}-\mathbf{\hat X}^{(i)}_{t'}\right\rVert_2^2
\end{equation}

\section{Experimental Results}


\paragraph{Implementation details.} We optimize our representation with Adam \cite{kingma2014adam}.
We use 18k gaussians for the foreground and 1.2M for the background. We fix the number of $\mathbb{SE}(3)$ motion bases to $28$ and obtain these from feature clustering (Sec. \ref{sec:feature_clustering}). For the depth alignment, we use points above the confidence threshold of 95\%. We show results on 7 10-sec long sequences at 30fps with a resolution of 512 $\times$ 288. Training takes about 30 minutes on a single NVIDIA A6000 GPU. Our rendering speed is about 30fps.

\vspace{-10pt}
\paragraph{Datasets.}
We conduct qualitative and numerical evaluation on Panoptic Studio \cite{joo2017panoptic} and a subset of Ego-Exo4D \cite{grauman2023egoexo4d} which we call ExoRecon. 

Panoptic Studio is a massively multi-view capture system which consists of 480 video streams of humans performing skilled activities. Out of these 480 views, we manually select 4 camera views, 90$^\circ$ apart to simulate the same exocentric camera setup as Ego-Exo4D. Given these 4 training view cameras, we find 4 other intermediate cameras 45$^\circ$ apart from the training views, and use these for evaluating novel view synthesis from 45$^\circ$ camera views.

For in-the-wild evaluation of sparse-view reconstruction, we repurpose Ego-Exo4D \cite{grauman2023egoexo4d}, which includes sparse-view videos of skilled human activities. While many Ego-Exo4D scenarios are out of scope for dynamic reconstruction with existing methods (due to fine-grained object motion, specular surfaces, or excessive scene clutter), we find one scene each from the 6 different scenarios in Ego-Exo4D with considerable object motion: \textit{dance}, \textit{sports}, \textit{bike repair}, \textit{cooking}, \textit{music}, \textit{healthcare}. For each scene, we extract 300 frames of synchronized RGB video streams, captured from 4 different cameras with known parameters. We remove fisheye distortions from all RGB videos and assume a simple pinhole camera model after undistortion. We call this subset ExoRecon, and show results on these sequences. Please see the appendix for more visuals.

\begin{table}[t]  
\centering
\resizebox{0.49\textwidth}{!}{%
\begin{tabular}{l|ccccc}
\toprule
\textbf{Method} & \textbf{PSNR}~$\uparrow$ & \textbf{SSIM}~$\uparrow$ & \textbf{LPIPS}~$\downarrow$ & \textbf{IOU}~$\uparrow$ & \textbf{AbsRel~($\downarrow$)}   \\
\midrule

SOM     & 16.73 & 0.554 & 0.491 & 0.287 & 0.578 \\
Dyn3D-GS     & 23.31 & 0.776 & 0.316  & --- &  0.273\\

MV-SOM      &  21.56 &  0.541 &  0.433 & 0.482  & 0.413\\
\textbf{MonoFusion} & \textbf{25.73} & \textbf{0.847} & \textbf{0.158} & \textbf{0.943} & \textbf{0.188} \\
\bottomrule
\end{tabular}
}
\caption{\textbf{Quantitative analysis of 45$^\circ$ novel-view synthesis on Panoptic Studio.} We benchmark our method against state-of-the-art approaches by evaluating both the dynamic foreground region and the entire scene. Notably, the evaluation is conducted on novel views where the cameras are at least $45^\circ$ apart from all training views. We additionally evaluate the geometric reconstruction quality with absolute relative (AbsRel) error in rendered depth.}
\label{tab:panoptic_45_deg_eval}
\end{table}

\begin{figure}[t]    \includegraphics[width=0.97\columnwidth]{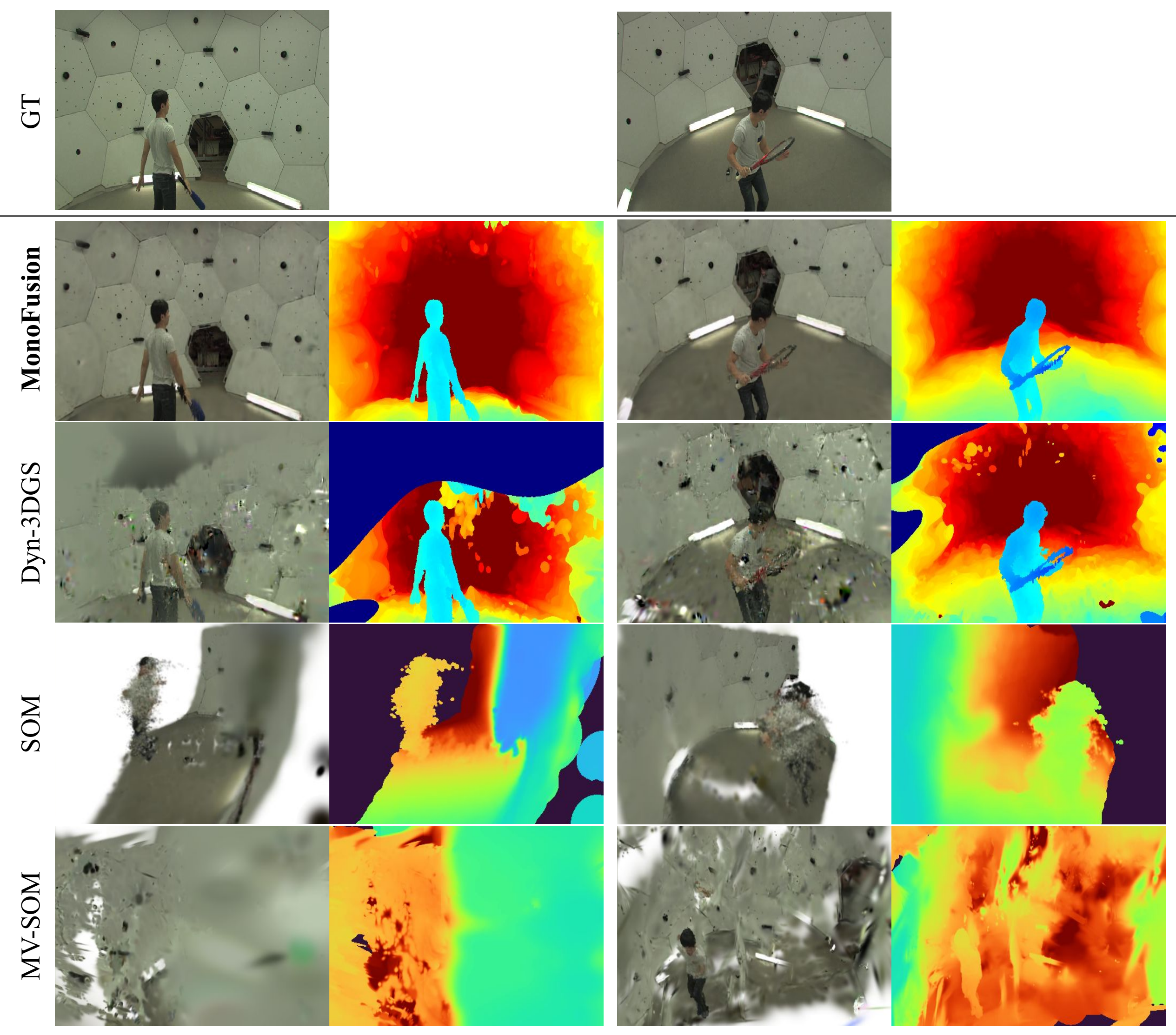}
    \captionof{figure}{\textbf{Qualitative results of $45^\circ$ novel-view synthesis results on Panoptic Studio.} We show qualitative novel-view synthesis results of our method compared to baselines on the softball (left) and tennis (right) sequences. We visualize the groundtruth RGB image for the $45^\circ$ at the top. Our rendered extreme novel-view RGB image closely matches ground truth. We find that all other baselines struggle to generalize to extreme novel views.
    }
    \label{fig:novel45degpanoptic}
    \vspace{-10pt}
\end{figure}

\begin{figure*}[t]
    \centering
     \includegraphics[width=0.97\textwidth]{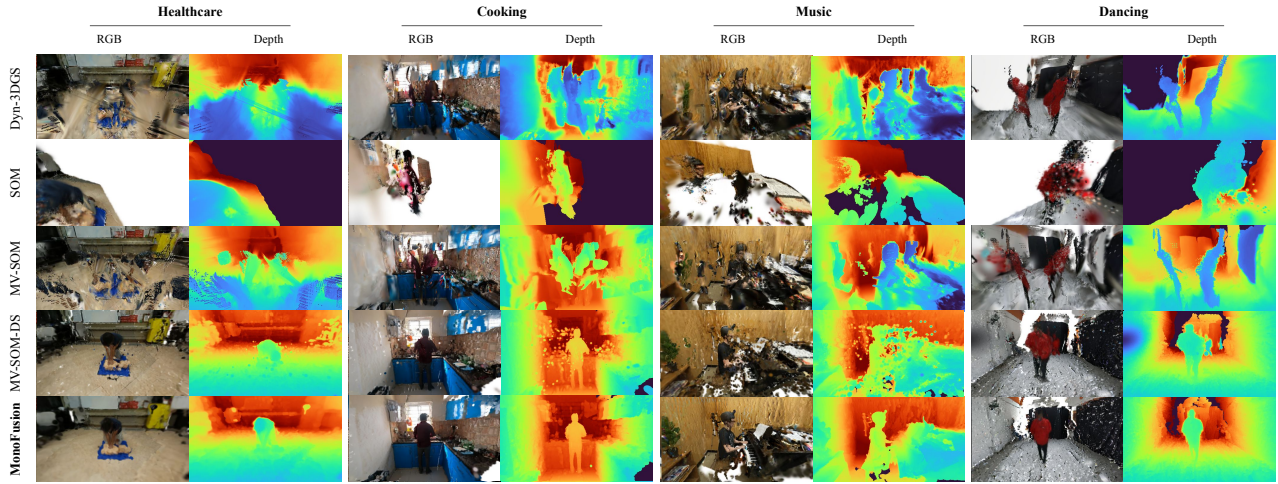}
    \captionof{figure}{\textbf{Qualitative results of 45$^\circ$ extreme novel view synthesis results on ExoRecon (1/2).} We visualize the rasterized RGB image and depth map from each method for 4 diverse EgoExo sequences. Existing monocular methods (Row 2, ``SOM'') and their extension to multi-view (Row 3, ``MV-SOM'') produce poor results rendered from a drastically different novel view. MV-SOM improves upon SOM by optimizing a 4D scene representation with four view constraints, but it still suffers from duplication artifacts. Our method's careful point cloud initialization and feature-based motion bases further improve on MV-SOM. Even after running MV-SOM with multi-view-consistent depth from DUSt3R (Row 4, ``MV-SOM-DS''), we find that it still fails due to reduced depth quality, often caused by suboptimal pairwise depth predictions on humans. Please see the appendix for more baseline comparisons: we find that multi-view diffusion methods contain additional hallucinations and imperfect alignment between different input views, and per-frame sparse-view 3D reconstruction methods suffer from temporal inconsistency, blurry reconstructions and missing details.}
    \label{fig:novel45degexo1}
\end{figure*}


\paragraph{Metrics.} We follow prior work \cite{luiten2023dynamic, yang2024storm} in evaluating the perceptual and geometric quality of our reconstructions using PSNR, SSIM, LPIPS and absolute relative (AbsRel) error in depth. We compute these metrics on the entire image, and also on only the foreground region of interest. We additionally evaluate the quality of the dynamic foreground silhouette by reporting mask IoU, computed as $(\mathbf{\hat M}\&\mathbf{M})/(\mathbf{\hat M}||\mathbf{M})$. Similar to prior work \cite{yang2024storm}, our evaluation views are a set of held-out frames, subsampled from the input videos from 4 exocentric cameras, in both Panoptic Studio and ExoRecon.

Note that since the cameras in our setup are stationary, above evaluation only analyses the {\em interpolation} quality of different methods. More explicitly, we also benchmark novel-view synthesis on Panoptic Studio with an evaluation camera placed 45$^\circ$ away from the training view cameras. Since such a ground-truth evaluation camera is not available in ExoRecon, we only show qualitative results.

\vspace{-10pt}
\paragraph{Baselines.}
We compare our method with prior work on dynamic scene reconstruction from single or multiple views. Among methods that operate on monocular videos, we run Shape of Motion \cite{wang2024som} on 8 scenes from Panoptic Studio following the setup of Dynamic 3D Gaussians \cite{luiten2023dynamic} and our curated dataset ExoRecon that covers 6 diverse scenes. Finally, we consider two multi-view dynamic reconstruction baselines, Dynamic 3D Gaussians \cite{luiten2023dynamic}, and a naive multi-view extension of Shape of Motion (MV-SOM). To construct the latter baseline, we simply concatenate the Gaussians and motion bases from four independently-initialized instances of single-view SOM, and optimize all four instances jointly. We also evaluate a variant of MV-SOM with globally-consistent depth (denoted MV-SOM-DS), obtained by running per-frame DUSt3R on the 4 input views and fixing camera poses to ground-truth during DUSt3R's global alignment. Despite using our same hyperparameters, MV-SOM-DS has more visual artifacts due to reduced depth quality, suggesting the importance of our DUSt3R+MoGe design. In the appendix, we verify that all baselines reconstruct reasonable training views. 

\subsection{Comparison to State-of-the-Art}

\noindent\textbf{Evaluation on held-out views.} In Tab. \ref{exp:ucsd_numerical}, we compare our method to recent dynamic scene reconstruction baselines \cite{zhang2024monst3r, luiten2023dynamic, wang2024som}, following evaluation protocols from prior work \cite{wang2024som, yang2024storm}. Our method beats prior art on both Panoptic Studio and ExoRecon (Fig. \ref{fig:noveltimesynthesis}) datasets, when evaluated on held-out views across photometric (PSNR, SSIM, LPIPS) and geometric error (AbsRel) metrics. Note that when initializing Dynamic 3DGS \cite{luiten2023dynamic} with 4 views we find that COLMAP fails, and so the point cloud initialization for this baseline is from a 27-view COLMAP optimization.

Interestingly, we find that although the monocular 4D reconstruction method Shape of Motion (SOM) \cite{wang2024som} often fails to output accurate metric depth, it is robust to a limited camera shift. We hypothesize that the foundational priors of Shape of Motion allow it to produce reasonable results in under-constrained scenarios, while test-time optimization methods, especially ones that do not always rely on data-driven priors \cite{luiten2023dynamic}, can more easily fall into local optima (e.g. those caused by poor initialization) which are difficult to optimize out of via rendering losses alone.

\begin{figure}[t]    \includegraphics[width=0.98\columnwidth]{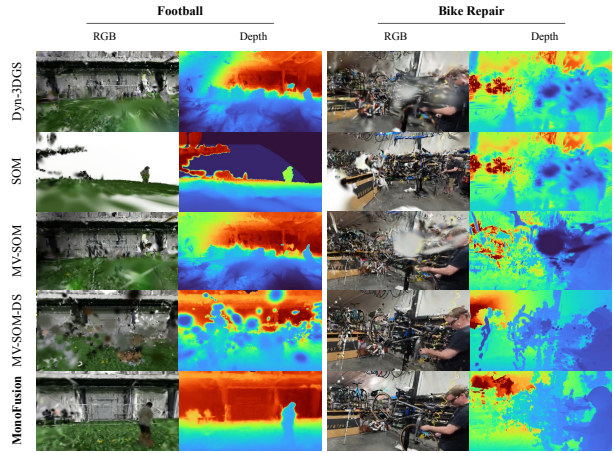}
    \captionof{figure}{\textbf{Qualitative results of 45$^\circ$ extreme novel view synthesis results on ExoRecon (2/2).} We show qualitative novel-view synthesis results of our method compared to baselines on challenging sequence on ExoRecon: highly-dynamic, large scene with small foreground \textit{football} (left) and complex, highly-occluded scene \textit{bike repair} (right). Notably MonoFusion significantly beats other baselines in terms of quality.  
    }
    \label{fig:novel45degexo2}
\end{figure}

\vspace{3pt}
\noindent\textbf{Evaluation on a 45$^\circ$ novel-view}. On Panoptic Studio, we use the four evaluation cameras (placed 45$^\circ$ apart from the training views) to evaluate our method's novel-view rendering capability. We also evaluate the novel-view rendered depth against a `pseudo-groundtruth' depth obtained from optimizing Dynamic 3DGS \cite{luiten2023dynamic} with all 24 training views. In Tab. \ref{tab:panoptic_45_deg_eval} and Fig. \ref{fig:novel45degpanoptic}, we find that our method outperforms all baselines, achieving state-of-the-art 45$^\circ$ novel-view synthesis. Qualitative results on ExoRecon are in Fig. \ref{fig:novel45degexo1} \& \ref{fig:novel45degexo2}.

\subsection{Ablation Study}
We ablate the design decisions in our pipeline in Tab. \ref{table:ablation_study}. Our proposed space-time consistent depth plays a crucial role in learning accurate scene geometry and appearance (yielding a 3.4 PSNR improvement, Row 1 vs 3). Next, we find that the feature-metric loss $\mathcal{L}_{\text{feat}} = \left\lVert\mathbf{\hat F}-\mathbf{F}\right\rVert$ provides a trade-off between learning photometric properties vs.learning foreground motion and silhouette. Although the PSNR decreases, we see an increase in mask IoU (Row 1 vs 2 and Row 3 vs 4). Freezing the color of all Gaussians across frames aids learning the motion mask, as measured by mask IoU. Finally, our motion bases from feature-clustering improve overall scene optimization (final row).

\begin{table}[t]
    \centering
    \resizebox{\columnwidth}{!}{
        \setlength{\tabcolsep}{1pt}
        \begin{tabular}{l|ccc|cccc}
        \toprule
        Method & 
        $\mathcal{L}_{\text{feat}}$ & 
        $\mathbf{d}_{n}$ &  
        $\mathbf{T}^{(b)}_{0\to t}$ & 
        $\uparrow$PSNR &
        $\uparrow$SSIM &
        $\downarrow$LPIPS &
        $\uparrow$IoU \\ \midrule
        Baseline & \xmark & \xmark & \xmark & 26.19 & 0.915 &  0.077 &  0.60\\
        + $\mathcal{L}_{\text{feat}}$ & \cmark & \xmark & \xmark & 25.39 & 0.933 & 0.087 & 0.63 \\
        + Our depth / no $\mathcal{L}_{\text{feat}}$  & \xmark & \cmark & \xmark & 29.55 & 0.944 & 0.037  & 0.73 \\
        + Our depth / $\mathcal{L}_{\text{feat}}$  & \cmark & \cmark & \xmark &29.31 & 0.941 & 0.041 &  0.75\\
        + Motion bases (\textbf{Ours}) & \cmark & \cmark & \cmark & \textbf{30.40}  & \textbf{0.947}  & \textbf{0.037} & \textbf{0.81}  \\
        \bottomrule
        \end{tabular}
    }
    \caption{\textbf{Ablation study of pipeline components.} We ablate our choice of feature-metric loss, spacetime consistent depth, and feature-based motion bases. While the proposed depth and feature-based motion bases considerably improve 4D reconstruction (evaluated by photometric errors), we find that our feature loss helps learn better motion masks (evaluated by IoU).
    }
    \label{table:ablation_study}
\end{table}

\begin{figure}[h]
    \centering
    \includegraphics[width=0.9\columnwidth]{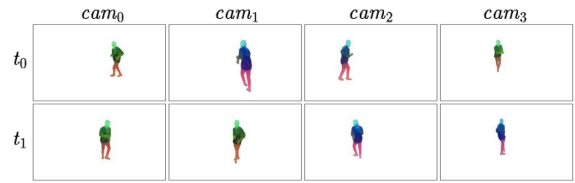}

    \captionof{figure}{\textbf{Spatial-Temporal Visualization of feature PCA.} We perform PCA analysis and transform the 32-dim features from \cref{sec:feature_clustering} down to 3 dimensions for visualization purposes. We find that the features are consistent across views and across time. Notably, when the person turns around between $t_0$ and $t_1$ in observations from $cam_1$ and $cam_2$, the feature remains robust and consistent. The semantic consistency of features aids explainability, provides a strong visual clue for tracking, and gives confidence in our feature-guided motion bases. 
    }

    \label{fig:short-a}
\end{figure}

\vspace{5pt}
\noindent\textbf{Velocity-based vs. feature-based motion bases} In the monocular setting, we empirically found that both designs performed equally well. However, in our 4 camera sparse view setting, we found that feature-based motion bases perform much better than velocity-based motion bases. The reason is that for velocity-based motion bases, we infer 3D velocity by querying the 2D tracking results plus depth per frame following Shape-of-Motion\cite{wang2024som}. Thus, noisy foreground depth estimates where the estimated depth of the person flickers between foreground and backward will negatively influence the quality of velocity-based motion bases, causing rigid body parts to move erratically. In contrast, feature-based motion bases, where features are initialized from more reliable image-level observations, are more robust to noisy 3D initialization and force semantically-similar parts to move in similar ways. To validate our points, in Fig. \ref{fig:short-a} we use PCA analysis to visualize the inferred features and find that they are consistent not only on temporal axis but also across cameras. 

\vspace{5pt}
\noindent\textbf{Effect of different number of motion bases.}
When the number of motion bases is not expressive enough (in our experience when the number of motion bases $<20$), there are often obvious flaws in the reconstruction, such as missing arms or the two legs joining together into a single leg. In reality, we do not observe that increasing the number of motion bases further hurts the performance. Empirically, the capacity of our design (which is \textbf{28} motion bases) can effectively handle different scene dynamics.

\section{Conclusion}



We address the problem of sparse-view 4D reconstruction of dynamic scenes. Existing multi-view 4D reconstruction methods are designed for dense multi-view setups (e.g. Panoptic Studio). In contrast, we aim to strike a balance between the ease and informativeness of multi-view data capture by reconstructing skilled human behaviors from four equidistant inward-facing static cameras. Our key insight is that carefully incorporating \textit{priors}, in the form of monocular depth and feature-based motion clustering, is crucial. Our empirical analysis shows that on challenging scenes with object dynamics, we achieve state-of-the-art performance on novel space-time synthesis compared to prior art.

\clearpage
\newpage 

{
    \small
    \bibliographystyle{ieeenat_fullname}
    \bibliography{main}
}

\clearpage
\setcounter{page}{1}
\appendix

\begin{center}
    \Large \textbf{Appendix}
\end{center}


\appendix

\section{Additional Baseline Comparisons}
To highlight the challenge of sparse-view 4D reconstruction, we compare with five additional baselines for monocular 4D reconstruction and sparse-view reconstruction. Our method remains state-of-the-art.

\begin{table}[htbp]
    \centering
    \resizebox{1\linewidth}{!}{%
    \begin{tabular}{llccc}
        \toprule
        \textbf{Method} & \textbf{PSNR} $\uparrow$ & \textbf{SSIM} $\uparrow$ & \textbf{LPIPS} $\downarrow$ & \textbf{Description}\\
        \midrule
        MV-MonST3R \cite{zhang2024monst3r} & 12.64 & 0.475 & 0.574 & Monocular 4D reconstruction\\
        MV-MegaSAM \cite{li2024megasam} & 14.25 & 0.578 & 0.412 & Monocular 4D reconstruction\\
        ViewCrafter \cite{yu2024viewcrafter} & 24.97 & 0.834 & 0.135 & Multi-view diffusion \\
        DNGaussian \cite{Li_2024_CVPR} & 27.34 & 0.897 & 0.103 & Sparse-view reconstruction \\
        InstantSplat \cite{fan2024instantsplat} & 29.11 & 0.909 & 0.082 & Sparse-view reconstruction\\
       MonoFusion & \textbf{30.64} & \textbf{0.930} & \textbf{0.055} & Ours\\
        \bottomrule
    \end{tabular}}
    \vspace{-14pt}
    \label{fig:quantitative_results}
\end{table}

\begin{figure*}[t]
    \centering
     \includegraphics[width=\textwidth]{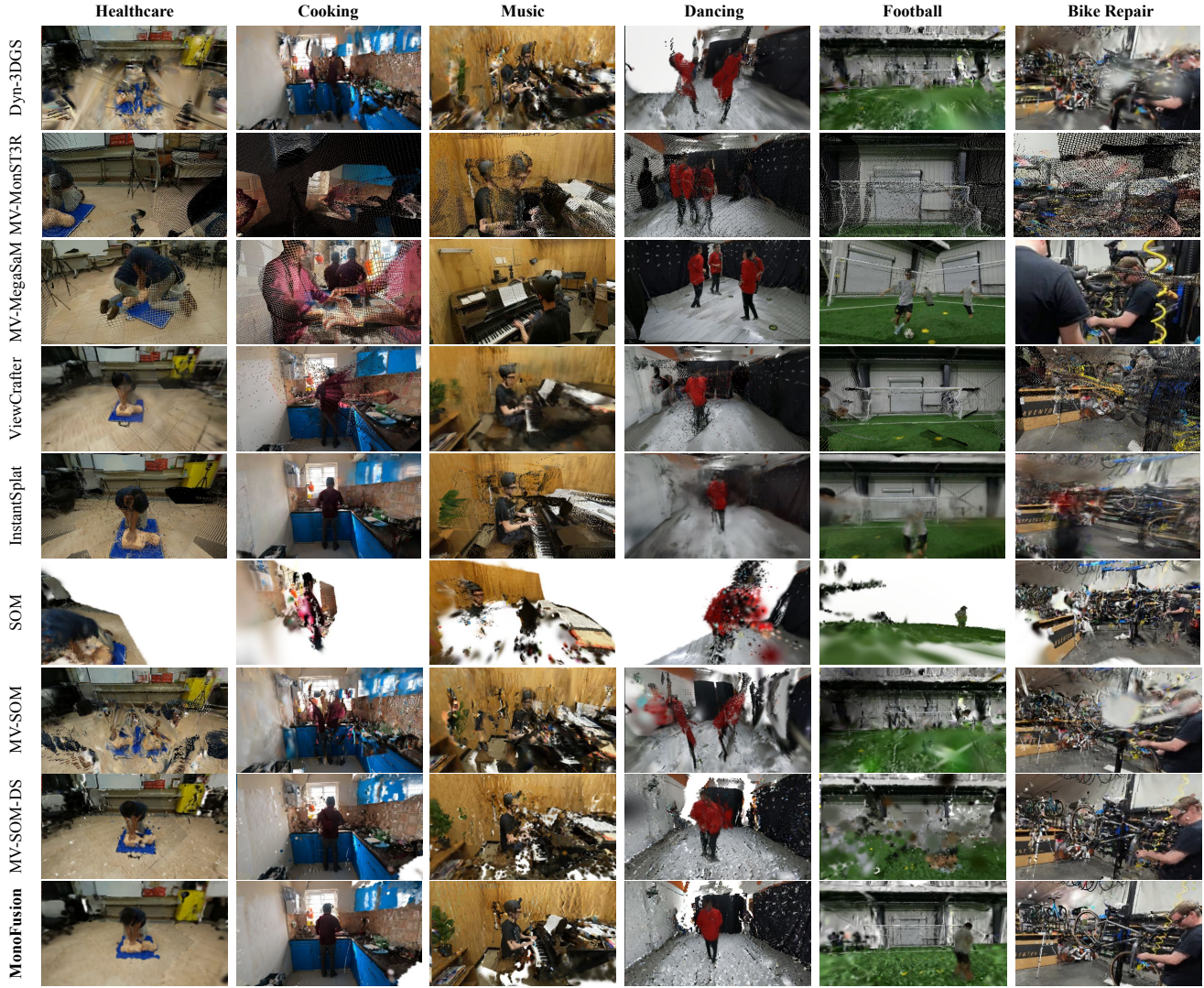}
    \captionof{figure}{\textbf{Qualitative results of 45$^\circ$ extreme novel view synthesis results on ExoRecon.} For each of the baseline methods considered in the main paper and in the appendix, we visualize the rasterized RGB image from each method for a 45$^\circ$ novel-view from one of the training views. Results suggest that prior work that is tuned for a dense camera setup struggles to work in the sparse-view case, and methods that are tuned for a monocular input cannot naively address a multi-view setup. Our proposed MonoFusion, successfully uses priors in the form of monocular depth and feature-based motion clustering to achieve the best of both worlds. 
    }
    \label{fig:qualitativeresults45exo}
\end{figure*}

\begin{figure*}[t]
    \centering
     \includegraphics[width=\textwidth]{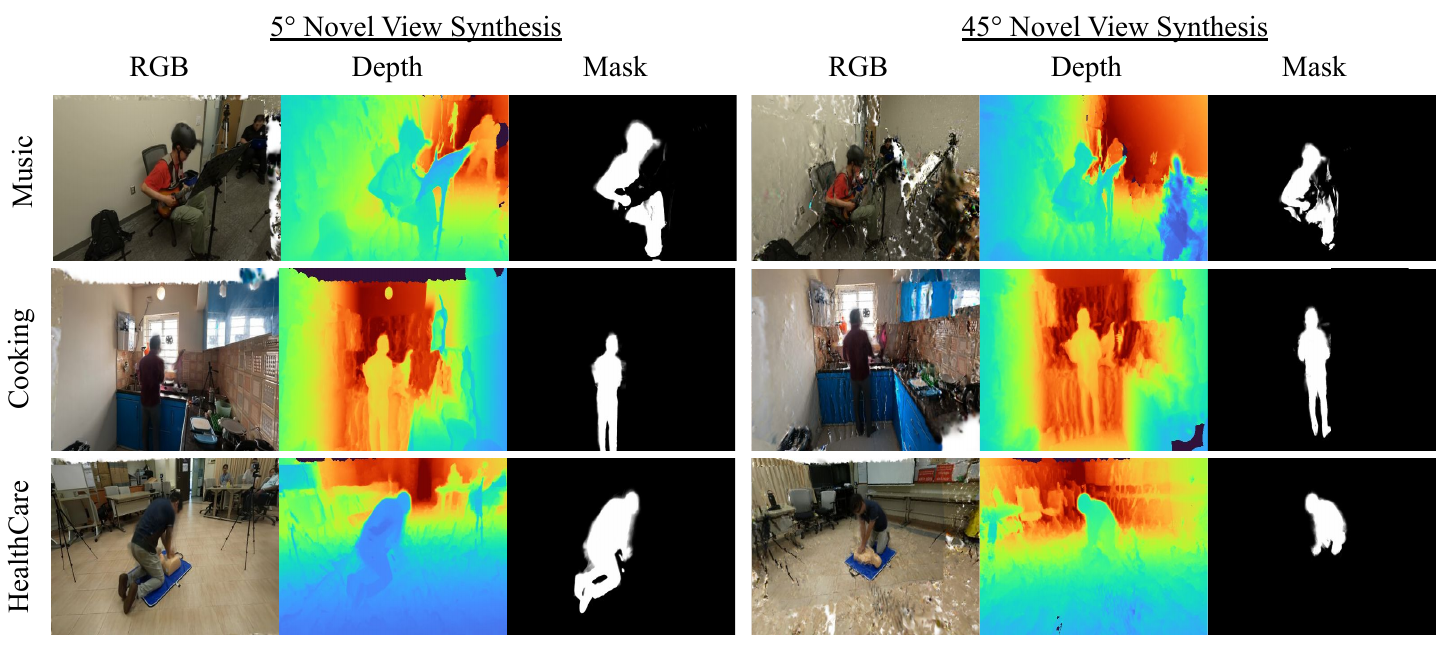}
    \captionof{figure}{\textbf{Novel view synthesis results from more video sequences.} In each row, we visualize the rasterized RGB image, depth map, and foreground mask from our method for various diverse scene including music (top), cooking (middle), and healthcare (bottom).  We include results for $5^\circ$ (left) and $45^\circ$ (right) novel view synthesis results. Notably, the rendered RGB and depth maps produce consistent reconstructions and plausible geometry.
    }
    \label{fig:qualitativeresults5and45}
\end{figure*}

\vspace{5pt}
\noindent\textbf{Baseline: Monocular 4D reconstruction}. We compare with MonST3R, as well as MegaSAM which claims better results than MonST3R. 
For each method, we report the best result among using a single-view video, concatenating the 4 views into longer video, or interleaving the 4 views (simulating a ``rotating" camera). MonST3R and MegaSAM fail in sparse-view scenarios, especially with large viewpoint shifts. When given concatenated or interleaved sparse-view videos as input, both models simply copy the input frames onto flat planes: notice the image corners in visual results. Given a single input video, both models are missing large regions due to occlusions and unseen geometry past the video boundary. Without non-trivially handling images from multiple wide-baseline input views, MonST3R and MegaSAM alone cannot produce view-consistent reconstructions.
\noindent\textbf{Baseline: Multi-view diffusion }. We use ViewCrafter, a video diffusion method for novel-view synthesis, to novel views at each timestep. ViewCrafter's rendered views are worse than our method's outputs, due to grid-like rendering artifacts, missing black regions of the scene, additional hallucinations, and imperfect alignment between different input views. \noindent\textbf{Baseline: Sparse-view methods }. We compare with alternative sparse-view methods InstantSplat and DNGaussian. Although they do not handle dynamic scenes, we run them independently per frame. Both methods suffer from blurry reconstructions and missing details, although InstantSplat benefits from cross-view consistency coming from DUSt3R. A qualitative analysis of all baselines for 5$^\circ$ and 45$^\circ$ novel view synthesis on the ExoRecon dataset is available in Fig. \ref{fig:qualitativeresults45exo} and \ref{fig:qualitativeresults5and45}.

\section{Training View Renderings}
\begin{figure}[t]
    \centering
     \includegraphics[width=0.49\textwidth]{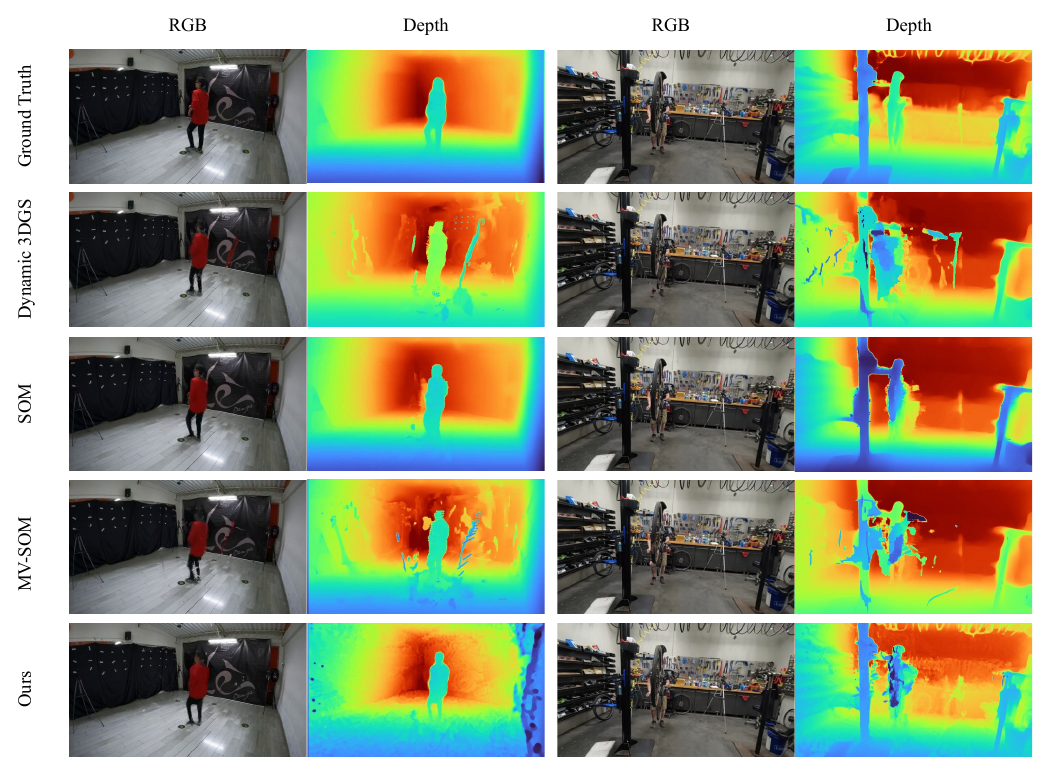}
    \captionof{figure}{\textbf{Training view results.} We visualize the rasterized RGB image and depth map from each method for the dancing (left) and bike repair (right) sequences. All methods are capable of producing reasonable training views and depth maps. It is worth noticing that in each iteration of optimization, we sample a batch of frames out of the video to optimize the overall loss. As the loss is optimized as a global minimum averaged over all frames, it is possible that some artifacts remain for certain frames. 
    }
    \label{fig:trainingview}
\end{figure}

To build confidence in our implementations, we validate every baseline we run by verifying that each method looks reasonable at training views (Fig. \ref{fig:trainingview}). It is worth noticing that in each iteration of optimization, we sample a batch of frames out of the video to optimize the overall loss. As the loss is optimized as a global minimum averaged over all frames, it is possible that some artifacts remain for certain frames. 

\section{Training Details}

In Tab. \ref{tab:lr_loss_row}, we report the learning rate and loss weights of Gaussians in our optimization process. These hyperparameters are shared across every scene that we evaluated on. Specifically, $\mathcal{L}_{\text{smooth\_bases}}$  
  enforces smooth motion bases by penalizing high accelerations in rotations and translations. $\mathcal{L}_{\text{smooth\_tracks}}$ promotes smooth object tracks by penalizing large accelerations in object positions across frames. $\mathcal{L}_{\text{depth\_grad}}$ aligns the gradients of the predicted and ground truth depth maps to preserve structural details. $\mathcal{L}_{\text{z\_accel}}$ penalizes high accelerations along the depth axis to reduce jitter in depth estimation. $\mathcal{L}_{\text{scale\_val}}$ constrains the variance of scale parameters of Gaussians to achieve consistent representations. We use Pypose\cite{zhan2023pypose} for held-out view synthesis.

\begin{table}[ht]
\centering
\caption{Training hyper-parameters: learning rates (left) and loss weights (right) shown row-by-row}
\label{tab:lr_loss_row}
\resizebox{0.49\textwidth}{!}{%
\begin{tabular}{lccc|lc}
\toprule
\multicolumn{4}{c|}{\textbf{Learning Rates}} & \multicolumn{2}{c}{\textbf{Loss Weights}}\\
\cmidrule(lr){1-4}\cmidrule(lr){5-6}
\textbf{Parameter} & \textbf{FG LR} & \textbf{BG LR} & \textbf{Motion LR} & \textbf{Loss Param.} & \textbf{Weight} \\
\midrule
means           & $1.6\times10^{-4}$ & $1.6\times10^{-4}$ & --                 & $w_{\text{rgb}}$            & $7.0$ \\
opacities       & $1\times10^{-2}$   & $1\times10^{-2}$   & --                 & $w_{\text{mask}}$           & $5.0$ \\
scales          & $5\times10^{-3}$   & $1\times10^{-3}$   & --                 & $w_{\text{feat}}$           & $7.0$ \\
quats           & $1\times10^{-3}$   & $1\times10^{-3}$   & --                 & $w_{\text{smooth\_bases}}$  & $0.1$ \\
colors          & \textbf{$0$}       & $1\times10^{-2}$   & --                 & $w_{\text{depth\_reg}}$     & $1.0$ \\
feats           & $1\times10^{-3}$   & $1\times10^{-3}$   & --                 & $w_{\text{smooth\_tracks}}$ & $2.0$ \\
motion\_coefs   & $1\times10^{-3}$   & --                 & --                 & $w_{\text{scale\_var}}$\footnotemark[1] & $0.01$ \\
rots            & --                 & --                 & $1.6\times10^{-4}$ & $w_{\text{z\_accel}}$       & $1.0$ \\
transls         & --                 & --                 & $1.6\times10^{-4}$ & $w_{\text{track}}$          & $2.0$ \\
\bottomrule
\end{tabular}}
\footnotetext[1]{Originally set to $0.01$.}
\end{table}

\section{Alternative Design Choice}
We explore higher-quality and metric-scale depth initialization, by replacing UniDepth in MV-SOM with improved metric depth estimators. Our results suggest that simply improving the quality of metric-scale depth in MV-SOM is not enough.
\begin{table}[h]  
\centering
\resizebox{0.49\textwidth}{!}{%
\begin{tabular}{l|ccccc}
\toprule
\textbf{Method} & \textbf{PSNR}~$\uparrow$ & \textbf{SSIM}~$\uparrow$ & \textbf{LPIPS}~$\downarrow$ & \textbf{IOU}~$\uparrow$ & \textbf{AbsRel~($\downarrow$)}   \\
\midrule
w/ UniDepth \cite{piccinelli2024unidepth}       & 26.91 & 0.890 & 0.138 & 0.845  & 0.474 \\
w/ Metric3Dv2 \cite{hu2024metric3dv2}     & 27.29 & 0.894 & 0.132 & 0.847 & 0.462 \\
w/ UniDepthV2 \cite{piccinelli2025unidepthv2universalmonocularmetric}   & 27.65 & 0.900 & 0.103 & 0.872 & 0.356 \\
\textbf{MonoFusion} & \textbf{30.64} & \textbf{0.930} & \textbf{0.055} & \textbf{0.963} & \textbf{0.290} \\
\bottomrule
\end{tabular}
}
\label{tab:ablation_metric_depth}
\end{table}

\newpage

\section{Limitations and Future Work}
\begin{figure}[h]
    \centering
    \includegraphics[width=\columnwidth]{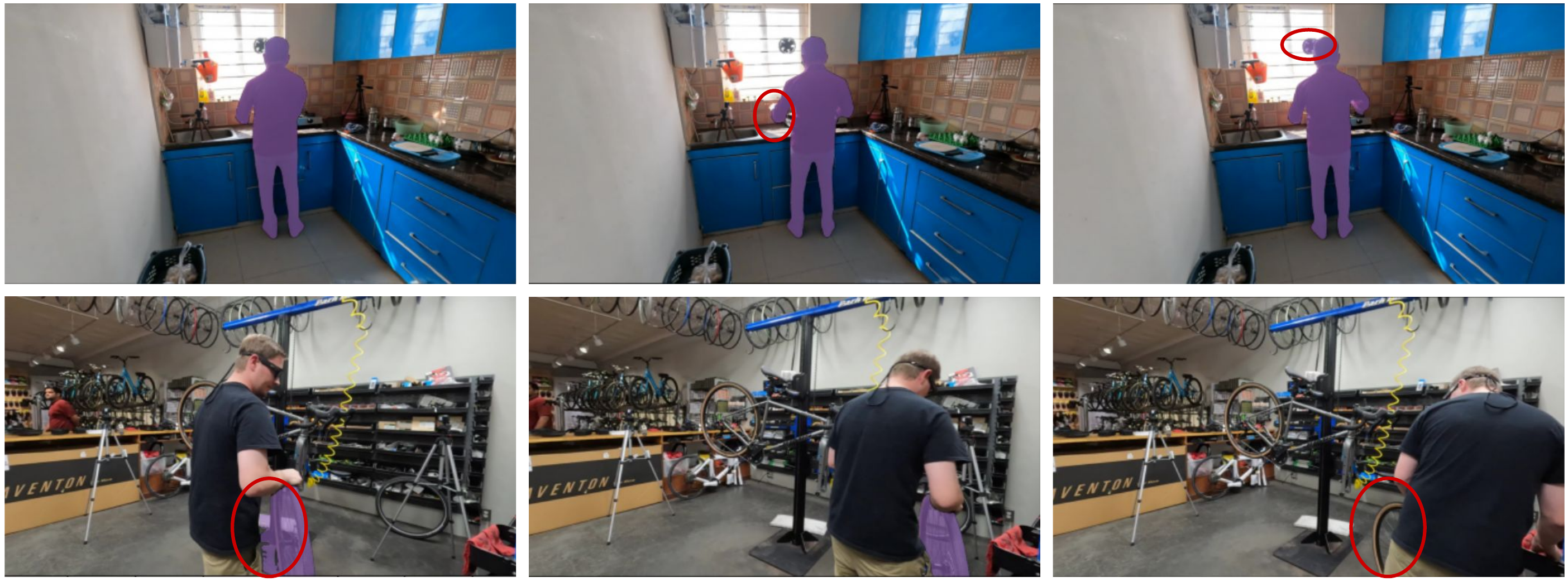}
    \captionof{figure}{\textbf{Failure example of SAM-V2.} We qualitatively inspect the SAM-V2 dynamic foreground masks on the kitchen (top) and bike repair (bottom) scenes. The dynamic mask is highlighted in \textit{purple}, and failures in dynamic mask estimation are highlighted in \textit{red circle}. We observe that SAM-V2 can miss important body parts (e.g. the person's hands) or get confused by the background (as shown in top row). Long-term occlusion will also lead to tracking failure (as shown in bottom row). These failure cases suggest that dynamic mask tracking in complex scenes remains an open challenge in works as in ~\cite{shan2021ptt, shan2022real, cui20213d, cui2019point, nie2024p2p, xu2024pillartrack, zhang2025yoloppa, sun2025yolov4svm, zeng2025janusvln}. It is also desired to have more fine-tuning on specific affordance as in~\cite{wan2024instructpart}. 
    }
    \label{fig:samfailure}
\end{figure}
We address two key limitations of our work. First, like previous methods, we rely heavily on 2D foundation models to estimate priors (e.g. depth and dynamic masks) for gradient-based differentiable rendering optimization. Thus, imprecise priors can harm the downstream rendering process. In addition, the current pipeline requires a user prompt to specify dynamic masks for each moving object \cite{cen2023segment, hu2024mvctrack}, which can be labor-intensive for complex scenes and may fail in some cases (Fig. \ref{fig:samfailure}). To solve this, distilling dynamic masks from foundation models or inferring dynamic masks from image level priors (as in \cite{zhang2024monst3r}) or completion\cite{yao2025depthssc} could be beneficial. 


\begin{figure}[h]
    \centering
    \includegraphics[width=\columnwidth]{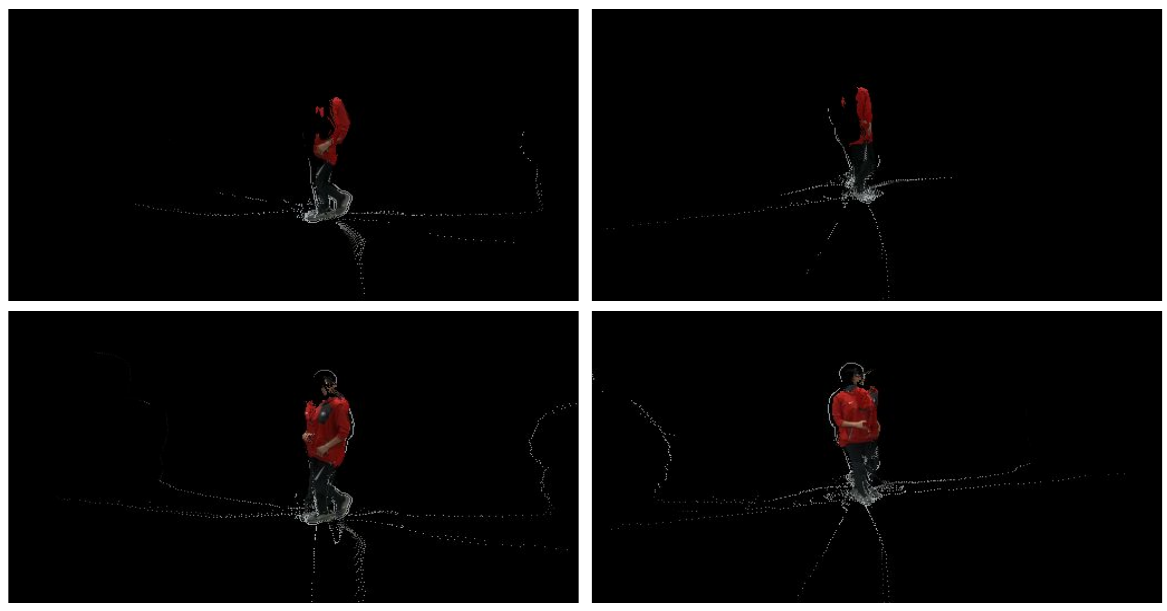}
    \captionof{figure}{\textbf{Visualization of foreground projection for different checkpoints.} Here we show the projection by known cameras and ground-truth foreground masks, using the point cloud from DUSt3R (\textit{top row}) and MonST3R (\textit{bottom row}) for two selected cameras (each column represents one camera). Notably, although MonST3R is fine-tuned on temporal frame sequences instead of multi-view information, MonST3R benefits from the presence of dynamic foreground movers in its fine-tuning dataset and thus gives a better foreground result.
    }
    \label{fig:ckp}
\end{figure}

Second, most off-the-shelf feed-forward depth estimation networks are trained on simple scene-level datasets, with few dynamic movers (e.g. people) in the foreground. In practice, we observe that the depth of humans in dynamic scenes is often incorrect when observed from other views (Fig. \ref{fig:ckp}). For example, DUSt3R often estimates the depth of a human to be the same as the depth of surrounding walls, causing the human to blend into the background. We believe that these fundamental problems with the depth predictions \textit{cannot} be solved by any alignment in the output space. To mitigate this issue, we plan to further fine-tune DUSt3R or MonST3R on existing dynamic human datasets.

\end{document}